\documentclass{article}
\usepackage{spconf,amsmath,graphicx}
\usepackage{CJKutf8}
\usepackage{bm}
\usepackage{adjustbox}
\usepackage{multirow}
\usepackage{booktabs}
\usepackage{cleveref}
\usepackage{spconf,amsmath,graphicx}
\usepackage{bm}
\usepackage{array,url,amssymb,bbding,cite}
\usepackage{spconf, amsmath, graphicx, setspace, epstopdf, booktabs}
\usepackage{tabularx, makecell, multirow, multicol, arydshln, stfloats}
\usepackage{cite}

\title{Matching-based Term Semantics Pre-training \\ for Spoken Patient Query Understanding}

\name{Zefa Hu$^{1,2}$, Xiuyi Chen$^{1,2}$, Haoran Wu$^{1,2}$, Minglun Han$^{1,2}$, Ziyi Ni$^{1,2}$, Jing Shi$^{2}$, Shuang Xu$^{2}$, Bo Xu$^{1,2}$
\thanks{This work is supported by the Key Programs of Chinese Academy of Sciences (No.ZDBS-SSW-JSC006-2), and the National Natural Science Foundation of China (No.62206294).}}
\address{
$^{1}$School of Artificial Intelligence, University of Chinese Academy of Sciences, Beijing, China\\
$^{2}$Institute of Automation, Chinese Academy of Sciences, Beijing, China\\
\small \tt \{huzefa2018, chenxiuyi2017, wuhaoran2018, hanminglun2018, niziyi2021\}@ia.ac.cn\\
\small \tt \{shijing2014, shuang.xu, xubo\}@ia.ac.cn}%

\begin{document}
\topmargin=0mm


%
\maketitle
\begin{abstract}
Medical Slot Filling (MSF) task aims to convert medical queries into structured information, playing an essential role in diagnosis dialogue systems. 
However, the lack of sufficient term semantics learning makes existing approaches hard to capture semantically identical but colloquial expressions of terms in medical conversations. 
In this work, we formalize MSF into a matching problem and propose a Term Semantics Pre-trained Matching Network (TSPMN) that takes both terms and queries as input to model their  semantic interaction. 
To learn term semantics better, we further design two self-supervised objectives, including Contrastive Term Discrimination (CTD) and Matching-based Mask Term  Modeling (MMTM). 
CTD determines whether it is the masked term in the dialogue for each given term, while MMTM directly predicts the masked ones.
Experimental results on two Chinese benchmarks show that TSPMN outperforms 
strong baselines, especially in few-shot settings\footnote{Our codes can be found at https://github.com/FlyingCat-fa/TSPMN.}. 
\end{abstract}
\section{Introduction}
\label{sec:intro}

Medical Slot Filling (MSF), which intends to automatically convert medical queries into structured information by detecting medical terms, has recently received increased attention 
\cite{liu2020meddg,DBLP:conf/aaai/ShiHCSLH20,shi2021understanding}.
It plays a vital role in diagnosis dialogue systems
\cite{DBLP:conf/acl/WeiLPTCHWD18, valizadeh2022ai} 
.
Different from conventional slot filling tasks in NLP that label the explicit words in a given utterance and extract the structured information (a.k.a slot-value pairs) based on the labeled words~\cite{DBLP:journals/taslp/MesnilDYBDHHHTY15, DBLP:conf/icassp/VuGAS16, DBLP:journals/corr/abs-1902-10909, DBLP:conf/icassp/QinLCKZ021}, 
MSF exists the non-alignment issue between a patient query and corresponding provided medical term slots \cite{DBLP:conf/aaai/ShiHCSLH20,  shi2021understanding, DBLP:conf/ecir/ZhouH11, DBLP:conf/coling/Rojas-BarahonaG16, DBLP:conf/acl/ZhaoF18}.
To be specific, 
colloquial expressions of terms in patient queries vary from formal expressions. 
As shown in Tabel \ref{tab:example}, the slot-value \texttt{Symptom:Bellyache} does not explicitly appear in any specific spans but is mentioned implicitly in the query.
Therefore, MSF requires a deeper understanding of term semantics with medical knowledge.  
Besides, medical data is more dependent on expert annotation, and the annotation is expensive to obtain in practice, which makes annotation data particularly insufficient.

\begin{table}
{
  \centering
    \begin{adjustbox}{scale=0.9,center}
    {\setlength{\tabcolsep}{10pt}
        \begin{tabular}{|l|}
            \hline
            Patient Query \\  
            \hline
            \multicolumn{1}{|p{20em}|}{My stomach feels bad these days, pain in the area above the navel, poop twice a day, belly bulge, shapeless, take cefixime currently, what happens?} \\
            \multicolumn{1}{|p{20em}|}{\begin{CJK*}{UTF8}{gbsn}我这几天肚子感觉难受，肚脐眼上面的位置疼痛，一天大便两次，肚子胀，不成型，目前在吃头孢克肟，这是怎么回事呢？\end{CJK*}} \\
            \hline
            Slot-values Pairs Label \\
            \hline
            Symptom:Bellyache (\begin{CJK*}{UTF8}{gbsn}腹痛\end{CJK*}) \\ 
            Symptom:Diarrhea (\begin{CJK*}{UTF8}{gbsn}稀便\end{CJK*}) \\
            Symptom:Abdominal Distension (\begin{CJK*}{UTF8}{gbsn}腹胀\end{CJK*}) \\
            Medicine:Cefixime (\begin{CJK*}{UTF8}{gbsn}头孢克肟\end{CJK*}) \\
            \hline
            \end{tabular}%
    }
    \end{adjustbox}
    \caption{An example of a patient query and the label that consists of slot-value pairs (e.g, Symptom:Diarrhea).}
    \vspace{-15pt}
  \label{tab:example}%
}
\end{table}%

Recent works~\cite{DBLP:conf/aaai/ShiHCSLH20,shi2021understanding, DBLP:journals/corr/abs-2203-09946} have been proposed to address the above problems, which can be generally grouped into two categories: multi-label classification and generative methods. 
The first category of methods~\cite{DBLP:conf/aaai/ShiHCSLH20,shi2021understanding} regards pre-defined slot-value pairs as different classes. 
They can utilize unlabeled patient queries with doctor responses to produce pseudo labels for data augmentation~\cite{meng2021mixspeech}.
However, it requires that unlabeled data map to the limited pre-defined terms,
making it challenging to exploit a larger unlabeled medical conversation corpus. 
The second category of methods~\cite{DBLP:journals/corr/abs-2203-09946}
commonly models MSF as a response generation task through a dialog prompt. 
In this way, MSF benefits from dialogue-style pre-training utilizing the large unlabeled medical dialogue corpus. 
However, the divergence between MSF and the response generation task inevitably undermines the performance. 
Besides, as the model generates terms in sequential order, the accumulated errors will be propagated to the later steps~\cite{DBLP:journals/corr/LiCLGC17}.

Unlike these approaches, we propose a Term Semantics Pre-trained Matching Network (TSPMN) that takes both terms and queries as input, 
leveraging the terms as task-related knowledge 
~\cite{DBLP:conf/icassp/HanDZX21, DBLP:conf/icassp/HanDLCZMX22}.
Therefore, the model only needs to learn how to match between the queries and given terms, which reduces the data restrictions.
Moreover, two self-supervised tasks are proposed for TSPMN to learn term semantics better, including Contrastive Term Discrimination (CTD) and Matching-based Mask Term  Modeling (MMTM). 
CTD is a matching task close to MSF, while MMTM is an adaptive Mask language Modeling (MLM) task to predict masked term tokens better by matching with golden tokens. 
In this way, TSPMN can not only use large-scale medical dialogue
corpora for pre-training, but also reduce the divergence between the pre-training and fine-tuning phases.
Experimental results on two Chinese benchmarks show that TSPMN outperforms strong baselines, especially in few-shot settings 
.

\begin{figure*}
\centering
\includegraphics[width=0.9\textwidth]{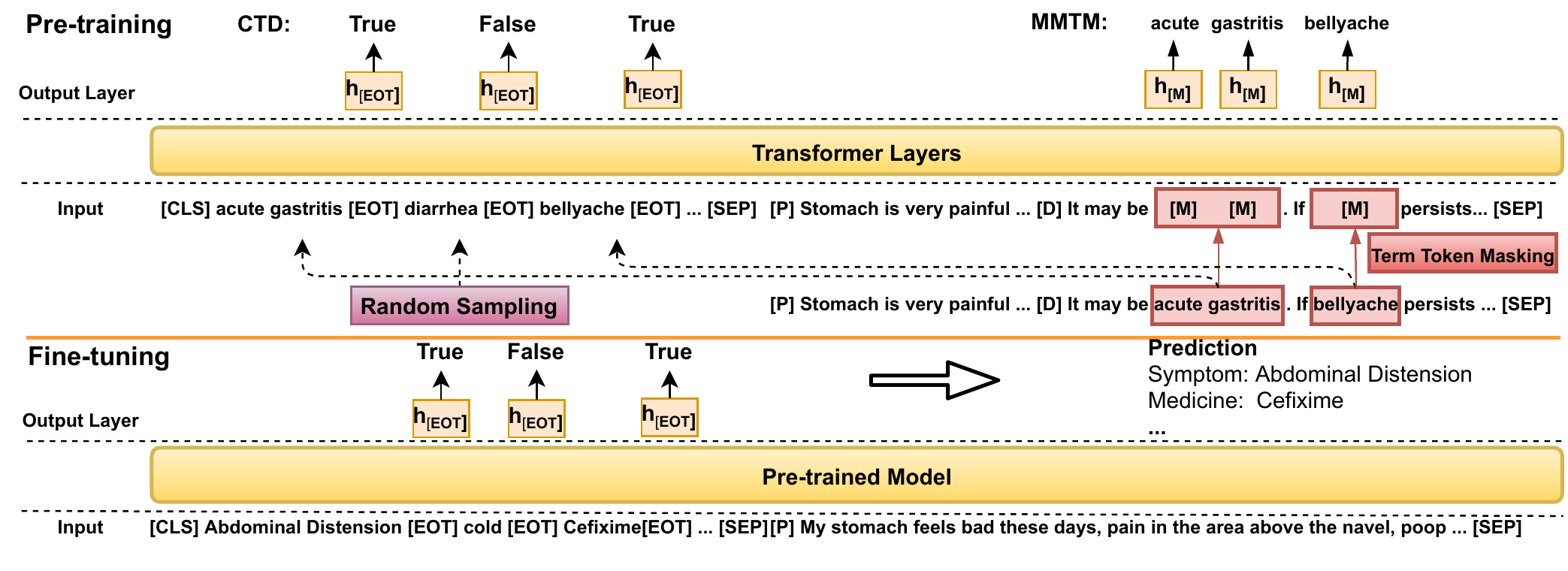}
\vspace{-5pt}
\caption{Illustration of Term Semantics Pre-trained Matching Network (TSPMN). 
The input consists of a term sequence and a medical dialogue/patient query.
\texttt{[EOT]} is the separator for the previous term.
\texttt{[P]} and \texttt{[D]} represents patient and doctor, respectively.
TSPMN first learns term semantics through our self-supervised tasks. 
Then the pre-trained TSPMN is fine-tuned to match candidate terms and the patient query for MSF. 
Note that all examples are translated from Chinese, and the example of fine-tuning is from Table~\ref{tab:example}.%
}
\label{TSPMN}
\vspace{-10pt}
\end{figure*}

\section{Problem Statement}
Given a patient query ${q}$ containing colloquial expressions, Medical Slot Filling (MSF) task aims at transforming the query ${q}$ into the grounded formal representation with discrete logical forms \texttt{(slot: value)}. The candidates of \texttt{slot} and \texttt{value} are pre-defined according to Medical Knowledge Graphs, where the value is a medical term, and the slot represents the category of the term (e.g., \texttt{(Symptom: Bellyache)}). We formulate MSF as a matching problem, in which we match each term candidate to the patient query ${q}$ to determine whether the candidate appears in ${q}$. 

\section{Approach}
This section presents how Term Semantics Pre-trained Matching Network (TSPMN) models Medical Slot Filling (MSF) by matching terms and patient queries. 
Then two term semantics pre-training tasks for TSPMN will be introduced.

\subsection{Matching for MSF}
\label{sec:Matching for MSF}
For efficient matching while considering the length limitations of the model input,
we first construct multiple term sequences by concatenating the terms in the term set $T$. 
Each term sequence is in the following form: 
\begin{equation}
    {S}_t = \left( {T_1}, \texttt{[EOT]}, \dots, {T_i}, \texttt{[EOT]}, \dots,  {T_n}, \texttt{[EOT]} \right),
\end{equation}
where $n$ and $\texttt{[EOT]}$ are the term number and the separator following each term, 
and $T_i$ represents the tokens of the $i$-th term.
Given a patient query, we concatenate each term sequence with the query as a whole sequence $x$, 
and encode ${x}$ with a pre-trained language model $\mathcal{H}$ such as BERT \cite{DBLP:conf/naacl/DevlinCLT19} to capture semantic information adequately:
\begin{equation}
    \left( \mathbf{h}_{\texttt{[CLS]}}, ... ,\mathbf{h}_{\texttt{[EOT]}}, ... ,\mathbf{h}_{\texttt{[SEP]}} \right)=\mathcal{H} \left( x \right).\label{con:encoder output}
\end{equation}
We take each hidden state $\mathbf{h}_{\texttt{[EOT]}}$ as the hidden state of the term before ${\texttt{[EOT]}}$. 
Then the probability about whether ${T_i}$ appears in the query is predicted as follows: 
\begin{equation}
    {p_i} \left( x;\theta \right) = \operatorname{Softmax} \left( \operatorname{FFN}\left( \mathbf{h}_{T_i}\right) \right), 
\label{con:modeling}
\end{equation}
where $\mathbf{h}_{T_i}$ means the hidden state of term $i$ and $\theta$ is the model parameter.
We map $\mathbf{h}_{T_i}$ to the scores of \texttt{True} and \texttt{False} independently through FFN, which means that term $i$ is mentioned in the query (\texttt{True}) or not (\texttt{False}), respectively. 
${p_i}(x;\theta) \in \mathbb{R}^{2}$ represents the scores normalized by softmax function.
If the normalized score of \texttt{True} is bigger than the \texttt{False}, we choose the term $i$ to fill the corresponding slot. 

The MSF loss function is defined as:
\begin{equation}
    \mathcal{L}_{MSF} = - \sum^{n}_{i=1} \sum_{k}{y}_{i,k} \log {p}_{i,k}, \quad {k}\in\{0,1\},\label{con:training}
\end{equation}
where $n$ and $k$ denote the number of terms and the index of \texttt{True} or \texttt{False}, and ${y}_{i} \in\{\left[1,0\right], \left[0,1\right]\}$ is the label indicating whether $T_i$ appears in the patient query.

\subsection{Term Sementics Pre-training}
Pre-trained language models (PrLMs) show excellent performance in many tasks
\cite{DBLP:conf/naacl/DevlinCLT19, meng2021offline}. 
There are also numerous PrLMs for dialogue representation
~\cite{mehri2019pretraining, DBLP:conf/acl/0001Z20, xu2021dialogue, he2022galaxy} 
or medical domain adaptation
~\cite{zhang2020conceptualized, wang2021pre, zhang2021smedbert}.
Inspired by these works, we focus on spoken understanding in medical dialogues and propose two self-supervised tasks to better model term semantics, which can not only use large-scale medical dialogue corpora, but also narrow the gap between those tasks and MSF as much as possible.
The details are described as follows.

\subsubsection{Matching for Pre-training}
We use three public unlabeled medical dialogue datasets MedDialog \cite{DBLP:conf/emnlp/ZengYJYWZZZDZFZ20}, KaMed \cite{DBLP:conf/sigir/LiRRCF0R21}, and ReMeDi-large\cite{yan2022remedi}, as pre-training corpora, which contain over 3.5M dialogues in more than 100 medical departments. 
The public sougoupinyin medical dictionary\footnote{\url{https://pinyin.sogou.com/dict/detail/index/15125}, updated to October 13, 2017.} and the medical dictionary THUOCL~\cite{han2016thuocl} are merged as a large medical terminology $T_{large}$. 
The terms in the knowledge base of the pre-training corpora are also added to $T_{large}$. 
Based on $T_{large}$, we retrieve the terms in each medical dialogue by string matching and construct dialogue-terms pairs for pre-training. 
Similar to the matching for MSF in \cref{sec:Matching for MSF}, we 
construct multiple term sequences by concatenating the terms
and the dialogue as the input.
Specifically, each term sequence consists of the sampled positive terms from the current dialogue and negative terms
that are not in the current dialogue.
The difference from MSF is that we only mask the sampled positive terms.
In this way, the model can learn term semantics from dialogue contexts
rather than just focus on string matching.
We denote the input as $x_{mask}$ and encode it in the same way as equation \ref{con:encoder output}:
\begin{equation}
    \left(\mathbf{h}_{\texttt{[CLS]}}, ... ,\mathbf{h}_{\texttt{[EOT]}}, ... ,\mathbf{h}_{\texttt{[M]}}, ... ,\mathbf{h}_{\texttt{[SEP]}}\right)=\mathcal{H}(x_{mask}),
\end{equation}
where $M$ means a masked token of terms.
The hidden states are used for our two self-supervised tasks.
The two tasks share the same input and encourage the pre-trained model to capture different aspects of semantics through multi-task learning. We define the total pre-training loss as the summation of two aforementioned losses:
\begin{equation}
    \mathcal{L}_{pretrain} = \lambda\mathcal{L}_{CTD} + (1-\lambda)\mathcal{L}_{MMTM}, 
\end{equation}
where $\lambda$ is a tunable weight used to adjust the contribution of different losses, $\mathcal{L}_{CTD}$ and $\mathcal{L}_{MMTM}$ are the losses of CTD and MMTM, respectively. The details of those two tasks are introduced in the following subsection.
\subsubsection{Self-supervised Tasks}
\label{sec:CTD}
\textbf{Contrastive Term Discrimination.} For each term in the term sequence, CTD aims to determine whether it belongs to the current dialogue. 
Similar to MSF in \cref{sec:Matching for MSF}, we use hidden states of $\texttt{[EOT]}$($\mathbf{h}_{T_i}=\mathbf{h}_{\texttt{[EOT]}_i}$) to represent the front term after matching with the patient query. 
The operation
is the same as equations \ref{con:modeling} and \ref{con:training}.
\\
\textbf{Matching-based Mask Term  Modeling.} 
This task is motivated by masked language modeling (MLM)~\cite{DBLP:conf/naacl/DevlinCLT19} with two improvements to match MSF:
1) 
MMTM only masks medical terms,
2) The masked always appears in the $T_{large}$.
Therefore, the model can not only learn semantics from the dialogue context but also more fully model the semantic interactions of the term and the dialogue. 
$\mathbf{h}_{\texttt{[M]}}$ is used to predict the mask. We compute the same cross-entropy loss as MLM.

\section{Experiments}

\subsection{Datasets and Evaluation Metrics}

\begin{table}[h]
{
  \centering
  \vspace{-5pt}
  \fontsize{9}{10}\selectfont 
    {\setlength{\tabcolsep}{6pt}
        \begin{tabular}{c|ccccc}
        \hline
        Dataset & Train & Dev   & Test  & Slot  & Value(Term) \\
        \hline
        MSL   & 1152  & 500   & 1000  & 1     & 29 \\
        MedDG  & 50965 & 6956  & 3645  & 4     & 155 \\
        \hline
        \end{tabular}%
    }
    \vspace{-5pt}
    \caption{Data statistics of MSL and MedDG datasets.}
  \label{tab:dataset}%
}
\vspace{-5pt}
\end{table}%

We evaluate our method on two Chinese medical datasets:
MSL~\cite{DBLP:conf/aaai/ShiHCSLH20} and MedDG \cite{liu2020meddg}.
MedDG was initially constructed for the medical dialogue system and labeled with the medical slots, which can be used for Medical Slot Filling (MSF). 
The statistics of the datasets are shown in Table \ref{tab:dataset}. 
For evaluation, we follow the MSL guidance~\cite{DBLP:conf/aaai/ShiHCSLH20} for all individual metrics: \textbf{Precision}, \textbf{Recall}, \textbf{Micro F1}, \textbf{Macro F1} and \textbf{Accuracy}.

\subsection{Implementation Details}
We initialize our model with Chinese BERT-base\cite{DBLP:conf/naacl/DevlinCLT19}. 
During the pre-training phase, 
the batch size is 48, and 1-bit Adam \cite{tang20211} is used as the optimizer.
We set the learning rate and pre-training epoch as $3\times 10^{-5}$ and $5$, respectively. 
And $\lambda$ is set to 0.9.
During the fine-tuning phase, AdamW \cite{DBLP:conf/iclr/LoshchilovH19} is used as our optimizer with an initial learning rate of $1\times 10^{-5}$. 
The batch size is 8 for MedDG and 32 for MSL. 
We set the term number $n$ of each term sequence to 20, 15 and 20 for pre-training, fine-tuning on MSL and MedDG, respectively.
\subsection{Main Results}
\label{sec:Results}

\begin{table*}[t]
\vspace{2pt}
  \centering\small
  \fontsize{9}{9}\selectfont 
      {\setlength{\tabcolsep}{9pt}
      \renewcommand{\arraystretch}{1.1}
        \scalebox{1}{\begin{tabular}{lcccccccccc}
        \toprule 
        \multicolumn{1}{c}{\multirow{2}{*}{\textbf{Model}}}&\multicolumn{5}{c}{\textbf{MSL}}&\multicolumn{5}{c}{\textbf{MedDG}}\\
        \cmidrule(r){2-6} \cmidrule(r){7-11}
    		& P & R & mi-F1 & ma-F1 & Acc & P & R & mi-F1 & ma-F1 & Acc\\
        \midrule
        \midrule DRNN~\cite{DBLP:conf/aaai/ShiHCSLH20}$\dagger$  & 83.43 & 67.85 & 74.83 & 65.17 & 52.5  & 96.55 & 97.34 & 96.95 & 82.8  & 73.39 \\
        DRNN+A~\cite{DBLP:conf/aaai/ShiHCSLH20}$\dagger$ & 82.11 & 70.86 & 76.07 & 67.42 & 51.9  & 98.53 & 96.69 & 97.6  & 83.62 & 75.25 \\
        DRNN+A+WS~\cite{DBLP:conf/aaai/ShiHCSLH20}$\ddagger$ & 82.94 & 79.44 & 81.15 & 76.95 & 58.3  &   -    &   -    &     -  &    -   & - \\
        TextCNN-Raw~\cite{shi2021understanding}$\dagger$ & 90.37 & 64.31 & 75.14 & 64.28 & 51.6  & 97.5  & 95.65 & 96.57 & 80.64 & 72.57 \\
        BERT-Raw~\cite{shi2021understanding}$\dagger$  & 89.78 & 88.63 & 89.2  & 87.03 & 70.9  & 99.3  & 99.38 & 99.34 & 84.82 & 74.15 \\
        BERT+TST~\cite{shi2021understanding}$\ddagger$ & 90.95 & 90.81 & 90.88 & 89.28 & 72.9  &    -   &   -    &   -    &   -    &  -\\
        PromptGen~\cite{DBLP:journals/corr/abs-2203-09946}$\mathsection$ & 89.11 & 87.57 & 88.34 & 87.75 & 79.6  &   -   &   -    &   -    &   -    &  -\\
        \midrule
        TSPMN-MedBERT & 90.91 & \textbf{91.11} & 91.01 & 90.45 & 81.4  & 99.38 & \textbf{99.5} & 99.44 & 86.51 & 98.44 \\
        TSPMN & \textbf{92.33} & 90.66 & \textbf{91.49} & \textbf{90.62} & \textbf{83.4} & \textbf{99.61} & 99.43 & \textbf{99.52} & \textbf{86.55} & \textbf{98.77} \\
        \quad  w/o MMTM & 91.14 & 90.66 & 90.90  & 89.74 & 82.80  & 99.45 & 99.43 & 99.44 & 86.34 & 98.44 \\
        \quad  w/o Pre-train & 85.76 & 88.40  & 87.06 & 86.36 & 74.50  & 99.21 & 99.28 & 99.25 & 86.00    & 97.86 \\
            \bottomrule
        \end{tabular}}%
    }
      \vspace{-3pt}
  \caption{Full training evaluation on MSL and MedDG datasets. $\dagger$: we cite the results of these models on MSL from the original papers~\cite{DBLP:conf/aaai/ShiHCSLH20, shi2021understanding}, and obtain the results on MedDG based on their released codes. 
    $\ddagger$: the models require homologous unlabeled data. 
    $\mathsection$: as the authors did not release their code, we cite the results of PromptGen on MSL from the original paper~\cite{DBLP:journals/corr/abs-2203-09946}. 
    }
  \label{tab:full_training}%
  \vspace{-5pt}
\end{table*}%

\begin{table*}[htbp]
  \centering\small
  \fontsize{9}{10}\selectfont 
  \resizebox{\linewidth}{!}
      {\setlength{\tabcolsep}{3pt}
      \renewcommand{\arraystretch}{1.1}
    \begin{tabular}{lccccccccc}
    \toprule
    \multicolumn{1}{c}{\multirow{2}[4]{*}{\textbf{Model}}} & \multicolumn{3}{c}{\textbf{1-shot}}            & \multicolumn{3}{c}{\textbf{2-shot}}            & \multicolumn{3}{c}{\textbf{5-shot}} \\
    \cmidrule(r){2-4} \cmidrule(r){5-7} \cmidrule{8-10}          & mi-F1 & ma-F1 & Acc & mi-F1 & ma-F1 & Acc & mi-F1 & ma-F1 & Acc \\
    \midrule
    \midrule DRNN+A & 20.66$\pm$3.32 & 11.63$\pm$3.15 & 7.16$\pm$1.88 & 33.69$\pm$3.38 & 27.96$\pm$4.89 & 11.74$\pm$1.15 & 52.83$\pm$4.84 & 50.24$\pm$5.3 & 23.74$\pm$3.56 \\
    DRNN+A+WS & 70.55$\pm$1.62 & 64.18$\pm$4.06 & 42.08$\pm$1.51 & 71.14$\pm$1.09 & 64.84$\pm$0.67 & 42.96$\pm$2.06 & 75.11$\pm$0.87 & 70.56$\pm$1.36 & 46.74$\pm$0.87 \\
    BERT-Raw  & 14.74$\pm$4.44 & 6.93$\pm$3.78 & 4.56$\pm$1.99 & 41.76$\pm$13.71 & 31.25$\pm$17.21 & 17.84$\pm$6.29 & 73.62$\pm$1.8 & 68.3$\pm$2.93 & 46.8$\pm$2.14 \\
    BERT+TST & 71.68$\pm$3.19 & 62.5$\pm$5.68 & 49.76$\pm$2.56 & 72.76$\pm$1.06 & 62.55$\pm$3.51 & 51.46$\pm$1.9 & 77.55$\pm$0.68 & 71.08$\pm$1.16 & 55.66$\pm$1.63 \\
    \midrule
    TSPMN-MedBERT & 78.19$\pm$0.86 & 77.81$\pm$0.25 & 53.32$\pm$1.6 & 79.34$\pm$0.93 & 79.1$\pm$1.43 & 55.76$\pm$2.14 & 83.34$\pm$1.04 & 83.27$\pm$1.13 & 64.08$\pm$1.96 \\
    TSPMN & \textbf{78.52$\pm$2.65} & \textbf{79.32$\pm$1.19} & \textbf{55.78$\pm$4.61} & \textbf{81.87$\pm$1.21} & \textbf{81.93$\pm$1.56} & 60.5$\pm$2.18 & \textbf{84.74$\pm$0.83} & 83.96$\pm$0.59 & \textbf{67.28$\pm$1.79} \\
    \quad  w/o MMTM & 77.82$\pm$2.32 & 78.24$\pm$1.47 & 55.68$\pm$3.63 & 81.37$\pm$1.2 & 81.49$\pm$0.59 & \textbf{61.2$\pm$1.86} & 84.22$\pm$0.83 & \textbf{84$\pm$0.91} & 66.08$\pm$1.84 \\
    \quad  w/o Pre-train & 75.12$\pm$1.3 & 74.69$\pm$0.77 & 48.78$\pm$2.45 & 77.59$\pm$0.45 & 76.87$\pm$0.48 & 53.86$\pm$0.98 & 81.61$\pm$1.64 & 81.64$\pm$1.46 & 60.78$\pm$3.13 \\
    \bottomrule
    \end{tabular}%
    }
\vspace{-8pt}
  \caption{Few-shot evaluation on MSL. The means and standard deviations over five runs are reported. }
  \label{tab:few_shot_MSF}%
\vspace{-10pt}
\end{table*}%

\textbf{Full Training Evaluation.} 
From Table~\ref{tab:full_training}
we can see that our model achieves new state-of-the-art results. 
The improvements of all metrics over baselines are statistically significant where $p < 0.05$ from significance testing. 
Compared with the classification method BERT+TST and the generative-based method PromptGen, 
which are enhanced or pre-trained on medical corpora, 
our TSPMN shows consistent improvements, which we attribute to TSPMN utilizing both large-scale unlabeled medical dialogue corpora and narrowing the discrepancy between the pre-training and fine-tuning phases. 
For further analysis, 
we remove the MMTM objective and then remove both MMTM and CTD, denoted as TSPMN w/o MMTM and TSPMN w/o Pre-train. 
We further replace our self-supervised objectives with the original BERT objectives, denote it as TSPMN-MedBERT. 
From the perspective of pre-training corpora, TSPMN and TSPMN-MedBERT perform better than TSPMN w/o Pre-train, illustrating the importance of continuous pre-training on large-scale medical dialogue corpora. 
From the perspective of pre-training tasks, the comparison of TSPMN and TSPMN-MedBERT indicates that the closer pre-training tasks get to the target task, the more performance gain can be achieved. 
The ablation experiments also verify the effectiveness of CTD and MMTM. 
\\
\textbf{Few-shot Evaluation.} 
We further evaluate our method in more challenging few-shot settings. 
In the $k$-shot setting, we select $k$ training examples from the original training set for each term to form the training dataset.
The initial validation and test data are still used for the few-shot evaluation. 
As shown in Table~\ref{tab:few_shot_MSF}, 
TSPMN achieves more performance gains than baselines in few-shot settings. 
Further, we find that TSPMN outperforms baselines on most measures even without pre-training, which validates the excellence of the novel matching paradigm of TSPMN in low resource scenarios. 
As shown in Table~\ref{tab:full_training} and Table~\ref{tab:few_shot_MSF}, compared with TSPMN-MedBERT, TSPMN achieves more relative improvement in few-shot settings than full training settings, which indicates that the smaller discrepancy between pre-training and fine-tuning phases is more significant in low resource scenarios.

\section{Conclusion}
The variation of terminology complexity between patients and formal providers requires a deeper and richer semantics understanding, which has been a headache in Medical Slot Filling (MSF) task. 
To learn term semantics thoroughly, this paper proposes Term Semantics Pre-trained Matching Network (TSPMN) with two self-supervised objectives, including Contrastive Term Discrimination (CTD) and Matching-based Mask Term  Modeling (MMTM). 
We reveal the excellence of TSPMN and the proposed training objectives through detailed experiments. 
The limitation of this paper is that the value of medical slots only consider as terms, which ignores the possible status corresponding to terms in more complicated scenarios, and we leave it to our future work.





\vfill\pagebreak

\bibliographystyle{IEEEbib}
\bibliography{strings_simpler}

\end{document}